\documentclass[letterpaper, 10 pt, conference]{ieeeconf}
\IEEEoverridecommandlockouts
\usepackage{amsmath}
\usepackage{cite}
\usepackage{bbold}
\usepackage{epsfig}
\usepackage[caption=false,listofformat=subsimple,labelformat=simple]{subfig}

\usepackage{multirow}
\usepackage{multicol}
\usepackage{cellspace}
\newcolumntype{P}[1]{>{\centering\arraybackslash}p{#1}}

\title{\LARGE \bf
3D Operation of Autonomous Excavator based on Reinforcement Learning through Independent Reward for Individual Joints
}
\author{Yoonkyu Yoo*, Donghwi Jung*, and Seong-Woo Kim%
\thanks{* indicates equal contribution.\newline \indent All authors are with Seoul National University, Seoul, South Korea.\newline
 {\tt\footnotesize \{farforaway,donghwijung,snwoo\}@snu.ac.kr}}}
\begin{document}
\maketitle
\thispagestyle{empty}
\pagestyle{empty}

\begin{abstract} 
In this paper, we propose a control algorithm based on reinforcement learning, employing independent rewards for each joint to control excavators in a 3D space. The aim of this research is to address the challenges associated with achieving precise control of excavators, which are extensively utilized in construction sites but prove challenging to control with precision due to their hydraulic structures. Traditional methods relied on operator expertise for precise excavator operation, occasionally resulting in safety accidents. Therefore, there have been endeavors to attain precise excavator control through equation-based control algorithms. However, these methods had the limitation of necessitating prior information related to physical values of the excavator, rendering them unsuitable for the diverse range of excavators used in the field. To overcome these limitations, we have explored reinforcement learning-based control methods that do not demand prior knowledge of specific equipment but instead utilize data to train models. Nevertheless, existing reinforcement learning-based methods overlooked cabin swing rotation and confined the bucket's workspace to a 2D plane. Control confined within such a limited area diminishes the applicability of the algorithm in construction sites. We address this issue by expanding the previous 2D plane workspace of the bucket operation into a 3D space, incorporating cabin swing rotation. By expanding the workspace into 3D, excavators can execute continuous operations without requiring human intervention. To accomplish this objective, distinct targets were established for each joint, facilitating the training of action values for each joint independently, regardless of the progress of other joint learning. To accurately assess performance, the model underwent training in simulation and was evaluated in simulation, after which the best-performing model was applied to a real excavator to verify if the excavator's bucket could effectively follow the target path. Ultimately, despite the expansion of the bucket's workspace into 3D, this model demonstrated rapid learning and achieved human-level performance on both linear and slope trajectories, surpassing human operators in certain tasks.
\end{abstract}
\section{Introduction}
\begin{figure}[t]
    \vspace{0.2cm}
    \centering
    \framebox{\parbox{0.48\textwidth}{\includegraphics[width=0.48\textwidth]{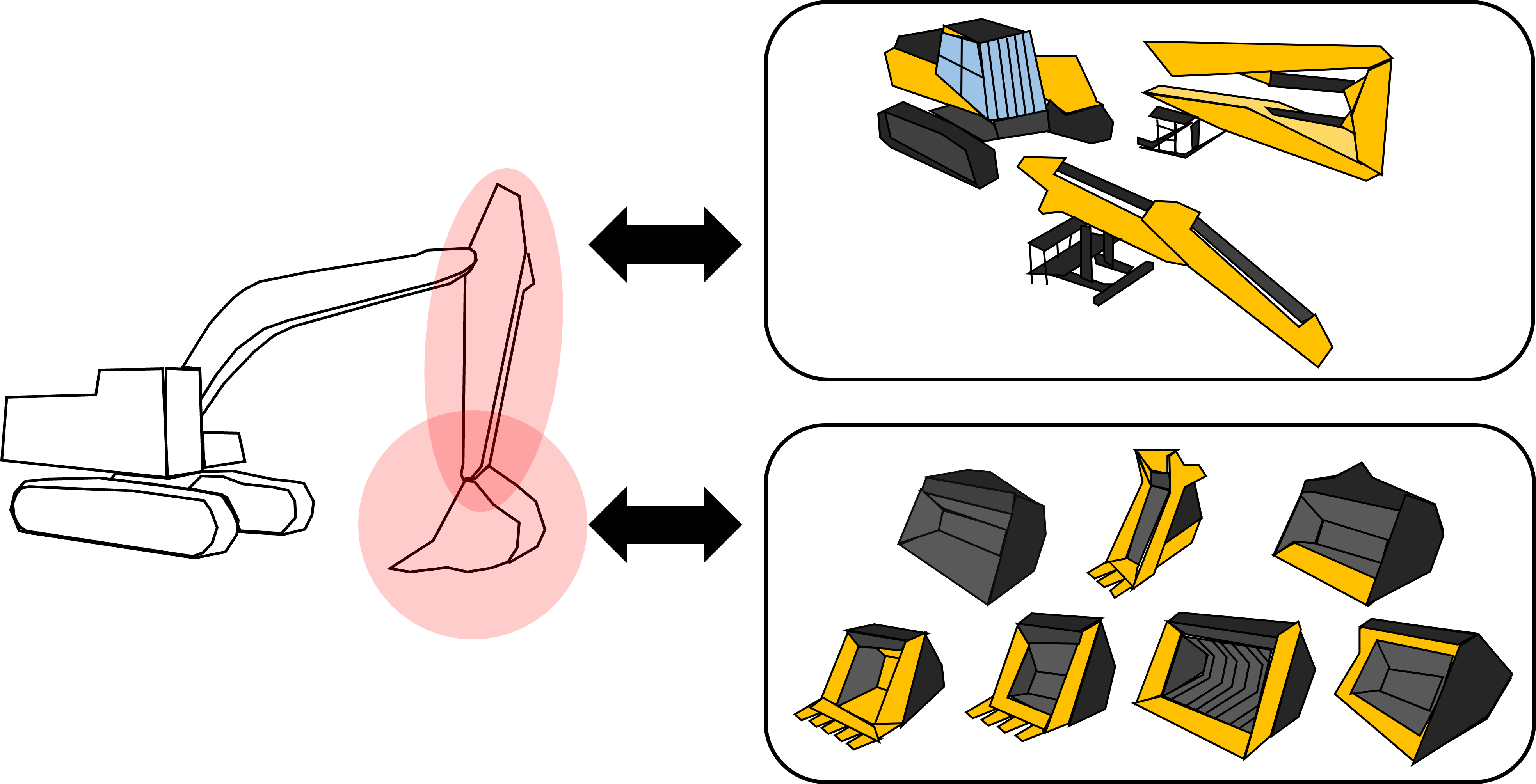}}}
    \caption{Examples of modifiable appearances in real-world scenarios that make it challenging to apply control algorithms to actual excavators.}
    \label{fig:custom}
        \vspace{-0.2cm}
\end{figure}
\begin{figure}[t]
    \vspace{0.2cm}
    \centering
    \framebox{\parbox{0.48\textwidth}{\includegraphics[width=0.48\textwidth]{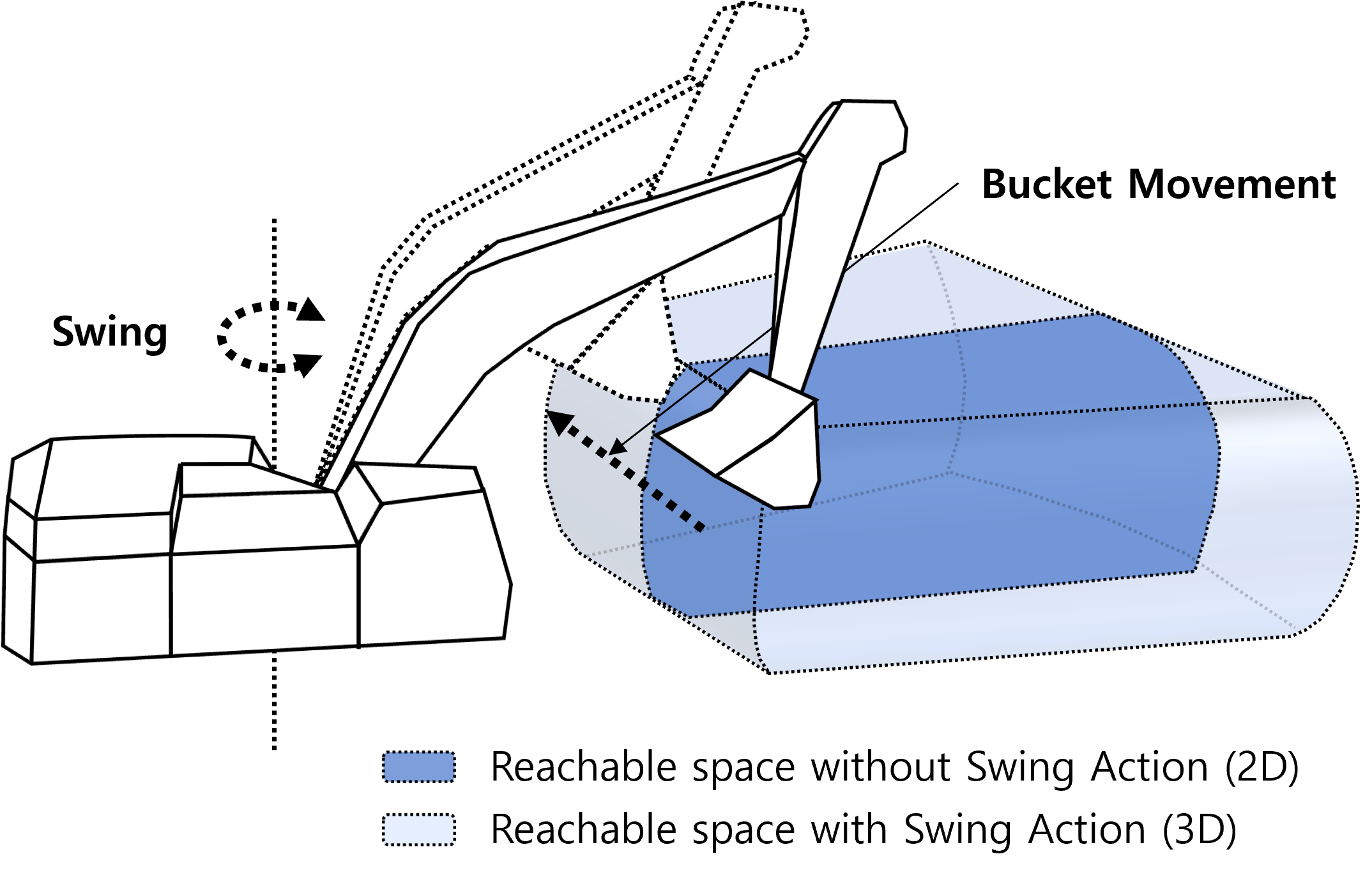}}}
    \caption{Expansion of bucket action space from 2D to 3D with cabin swing rotation that enables the excavator to perform a wider range of construction tasks.}
    \label{fig:bucket_space}
        \vspace{-0.2cm}
\end{figure}
\begin{figure*}[t]
    \vspace{0.2cm}
    \centering
    \framebox{\parbox{\textwidth}{\includegraphics[width=\textwidth]{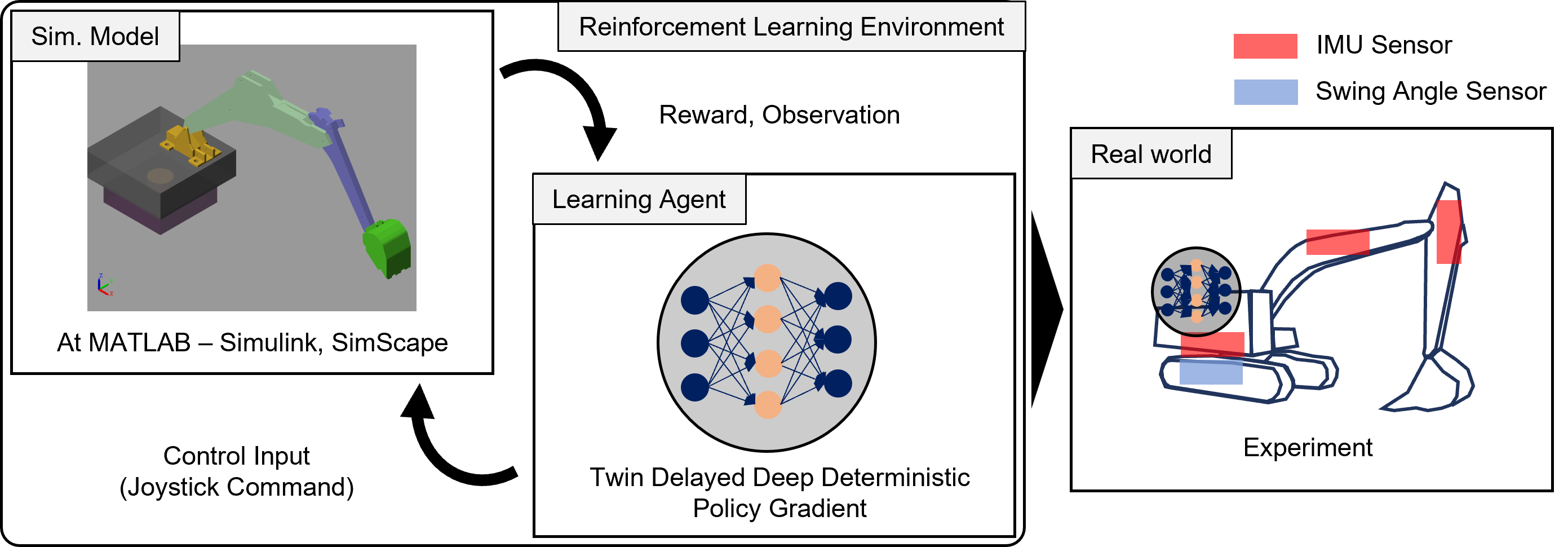}}}
    \caption{Process of our proposed method.}
    \label{fig:process}
        \vspace{-0.4cm}
\end{figure*}

At construction sites, there have been instances of safety accidents caused by errors made by heavy equipment operators. Consequently, efforts have been made to enhance safety in excavator operations by introducing unmanned technology to construction sites. Previous papers \cite{chi2012image,kim2018interaction,feng2018robotic,lee2021real,bender2015nonlinear,lu2022excavation,hodel2018learning, egli2020towards,egli2022general, egli2022soil} have not only focused on on-site sensing technologies for identifying workplace hazards\cite{chi2012image,kim2018interaction}, but also on the direct control of construction equipment\cite{feng2018robotic,lee2021real,bender2015nonlinear,lu2022excavation,hodel2018learning, egli2020towards,egli2022general, egli2022soil}. One example of unmanned control for construction equipment is in the case of excavators, which are used for digging the ground and moving soil or rocks. Currently, the majority of excavators are designed with hydraulic systems, leading to significant non-linearities in their control. As a result, the operator's skill plays a crucial role in the performance of excavator operations.

As explored in \cite{feng2018robotic,lee2021real,bender2015nonlinear}, there have been efforts to develop technologies that facilitate the operator's control of excavators by employing traditional control methods like PID (Proportional Integral Derivative) or MPC (Model Predictive Control). These methods offer the advantage of precision since they are control-based. However, as illustrated in Fig. \ref{fig:custom}, in real-world working conditions, the excavator's behavior often varies due to factors such as specific task requirements and environmental conditions. Consequently, for practical application in the field, algorithms capable of swiftly adapting to diverse environments become essential.\\

\indent Reinforcement learning is a machine learning approach widely utilized in control algorithms\cite{lu2022excavation,hodel2018learning, egli2020towards,egli2022general, egli2022soil}, wherein an agent employs a reward function to obtain appropriate rewards based on its actions and subsequently updates its model. Unlike conventional control methods mentioned in previous studies \cite{feng2018robotic,lee2021real,bender2015nonlinear}, these updates do not require prior knowledge of the mechanical model. This characteristic allows the model to adapt proactively to various excavator variations, as illustrated in Fig. \ref{fig:custom}. Consequently, in this paper, we introduce reinforcement learning to excavator scenarios. To employ reinforcement learning for model training, interaction data from various environments are necessary. However, conducting multiple tests with real excavators poses safety challenges. Therefore, as shown in Fig. \ref{fig:environments}, we constructed a simulation model identical to the real scenario and collected data within this simulated environment to facilitate the training of the reinforcement learning model. Subsequently, we applied this trained model to the actual excavator environment for performance validation.\\
\indent Unlike previous studies \cite{feng2018robotic,lee2021real,bender2015nonlinear,lu2022excavation,hodel2018learning,egli2020towards,egli2022general, egli2022soil}, our excavator model includes an additional cabin swing joint. As depicted in Fig. \ref{fig:bucket_space}, the addition of this swing joint enables the bucket to move in a 3D space rather than being restricted to a simple 2D plane, expanding its capabilities for a wider range of tasks. Consequently, the potential for practical applications on construction sites increases significantly. However, this expansion of the action space introduces challenges when learning a single target. Specifically, in the early stages of learning, the reliance on the cabin swing joint for bucket positioning can impede or slow down the learning process. To foster a more stable model learning process, we establish distinct learning targets for each joint and structure rewards accordingly. Furthermore, by introducing slope trajectories into the target tasks, a consideration not addressed in previous studies \cite{hodel2018learning,egli2020towards,egli2022general}, we assessed the excavator's ability to handle a broader range of tasks. Ultimately, this enhances the model's practical applicability.\\
\indent To the best of our knowledge, this is the first paper to incorporate the cabin swing joint of an excavator as an action target, enabling the bucket to move in a 3D space that more closely simulates real construction work environments compared to previous papers. To accomplish this, we have established separate learning targets for each joint and formulated independent rewards, which are then utilized within the reinforcement learning model.\\
\indent The contributions of our paper are as follows:
\begin{itemize}
    \item \emph{3D Action Space}: We extend the range of the bucket's movement from the traditional 2D plane to a 3D space by including cabin swing rotation as part of the action.
    \item \emph{Independent Reward}: In 3D action space, individual rewards that consider the characteristics of each joint exhibit superior performance when compared to a single distance-based reward.
    \item \emph{Simulation to Real World}: We illustrate that the model trained through simulation exhibits high accuracy in real excavator operations.
\end{itemize}
\section{Related Works}
In the context of conventional excavator control, commonly employed methods include model-free approaches such as PID \cite{feng2018robotic}, as well as model predictive methods like MPC \cite{lee2021real, bender2015nonlinear}. These approaches calculated the excavator's movements using formulas, offering the advantage of high accuracy in specific situations. However, as depicted in Fig. \ref{fig:custom}, if the excavator undergoes changes or if the surrounding environment changes, neglecting to integrate these modifications into the existing formulas leads to reduced accuracy. Furthermore, quantifying and directly integrating these changes into the formulas can be challenging, presenting a limitation. Separate from traditional control methods, learning-based approaches have also found applications in excavator control. Among these, there is literature \cite{lu2022excavation} that calculated the target position through reinforcement learning in camera sensor coordinates and operated it by computing control values for each joint using inverse kinematics. In their work, the environment was observed with a camera mounted at the top, and a neural network was employed to deduce the geometric representation of the object and calculate the target position in camera coordinates. Consequently, this method determined the angles of movement of the robotic arm using inverse kinematics and applied the computed values as control inputs. While this method exhibits high accuracy in calculating control values for points using kinematics, it requires additional sensors like cameras. Furthermore, it has the drawback of needing to measure and incorporate changes in the physical values of the excavator when they occur.\\
\indent To overcome these limitations, there is literature that employs an end-to-end approach utilizing reinforcement learning \cite{hodel2018learning, egli2020towards, egli2022general, egli2022soil} to directly derive control values for each joint. In \cite{hodel2018learning}, the authors addressed the fundamental and widely used task of linear bucket leveling in excavators. However, this paper conducted experiments solely through simulations and did not test the algorithm on real excavators. In contrast, papers such as \cite{egli2020towards, egli2022general, egli2022soil} utilized reinforcement learning models for excavator control, assessing their performance not only in simulations but also on real excavators. In \cite{egli2020towards, egli2022general}, by employing 3-4 degrees of freedom (DOFs) within a 2D plane, they examined the bucket's tracking performance for linear, circular, and grading trajectories. Additionally, Egli \emph{et al.} \cite{egli2022soil} proposed dynamic response methods for situations involving changes in soil viscosity. While the specific tasks targeted in these papers vary, a common feature is that the output of reinforcement learning directly controls the excavator using control values for joints, without the need for separate formulas such as inverse kinematics. However, these existing end-to-end approaches are limited by having a 2D-restricted action space for the bucket. In real construction scenarios, excavators manipulate buckets in 3D by utilizing cabin rotation. Recognizing this, we present a 4 DOFs excavator model in this paper, which includes the cabin's swing joint. Moreover, given that the expansion of the bucket's action space can make learning challenging when using distance-based rewards from a single target, we train the model using independent rewards calculated by setting separate targets for each joint.
\begin{figure}[t]
    \vspace{0.2cm}
    \centering
    \framebox{\parbox{0.48\textwidth}{\includegraphics[width=0.48\textwidth]{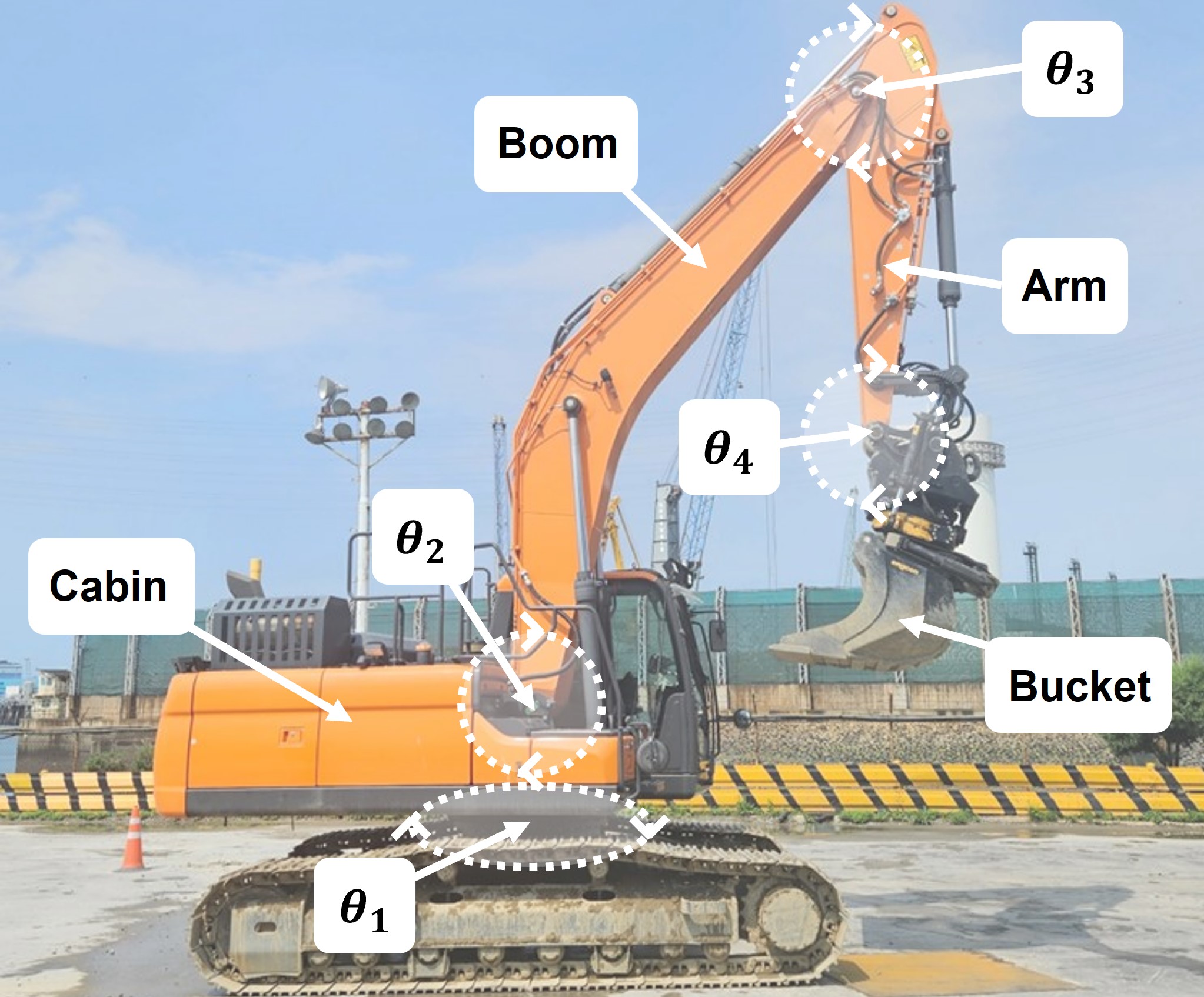}}}
    \caption{The kinematic architecture of excavator. Our target model contains four controllable joints.}
    \label{fig:real_excavator_w_instruction}
    \vspace{-0.4cm}
\end{figure}
\begin{figure*}[t]
    \vspace{0.2cm}
    \centering
    \framebox{\parbox{\textwidth}{\includegraphics[width=\textwidth]{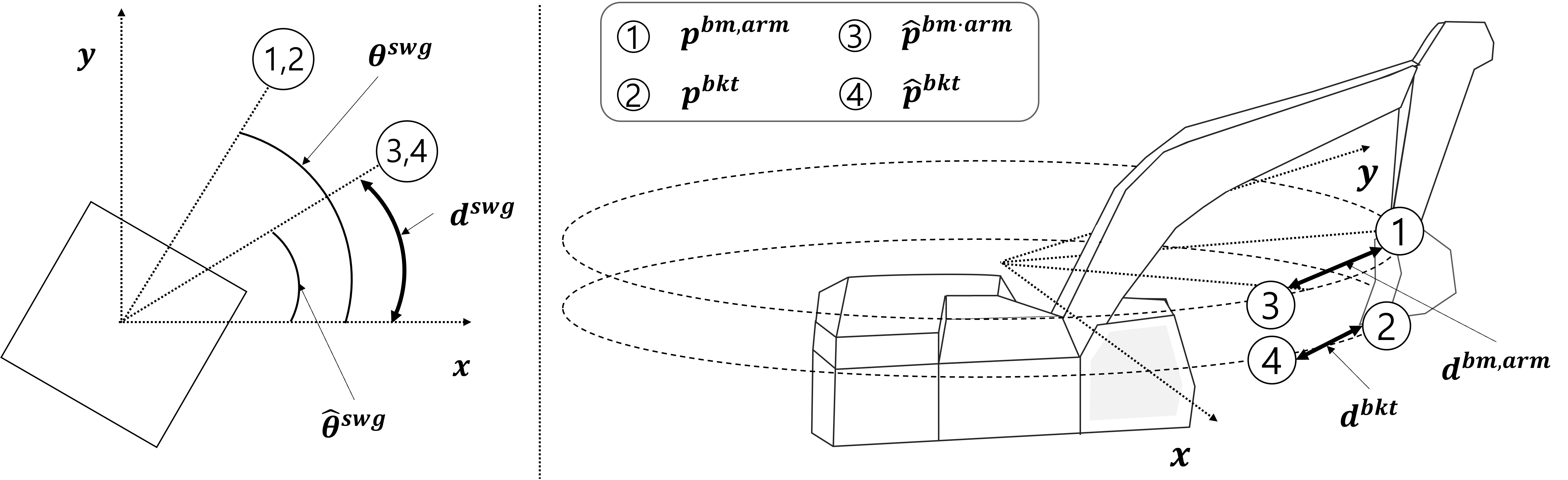}}}
    \caption{Partial rewards for joints of an excavator. Left: reward for a swing joint (top-view), Right: rewards for boom, arm, bucket joints (side-view).}
    \label{fig:rewards}
    \vspace{-0.2cm}
\end{figure*}
\begin{table}[t]
    \vspace{0.2cm}
\caption{Network design of TD3 layers\\(In: Input Size, H: Hidden Layer, Out: Output Size)}
\label{table:network_architecture}
\begin{center}
\renewcommand{\arraystretch}{2}

    \begin{tabular}{P{0.09\textwidth} P{0.05\textwidth} P{0.05\textwidth} P{0.05\textwidth} P{0.05\textwidth} P{0.05\textwidth}}\hline
    &In&H1&H2&H3&Out\\\hline
    Actor&17&180&180&180&4\\\hline
    Critic 1&17&180&180&180&4\\\hline
    Critic 2&17&64&180&180&4\\\hline
    \end{tabular}
    \vspace{-0.4cm}

\end{center}
\end{table}
\section{Methods}

\subsection{Environment}
The schematic diagram of the excavator used in this experiment is depicted in Fig. \ref{fig:real_excavator_w_instruction}. The excavator model consists of a total of 4 joints (swing, boom, arm, bucket), thus providing the reinforcement learning agent with a total of 4 DOFs. In this context, to define the joint action values in reinforcement learning, the joint velocities transmitted to each joint are denoted as $a_1, a_2, a_3$, and $a_4$. Based on these values, the reinforcement learning environment is formulated as follows:
\begin{gather}
    P = \{p_x,p_y,p_z\},\\
    \hat{P} = \{\hat{p}_x,\hat{p}_y,\hat{p}_z\},\\
    U = \{u_1,u_2,u_3,u_4\},\\
    S = P \cup \hat{P} \cup U,\\
    A = \{a_1,a_2,a_3,a_4\},\\
    R = \{r\},
\end{gather}
where $P$ represents the position of the bucket's end-tip, consisting of $p_x$, $p_y$, and $p_z$. Similarly, $\hat{P}$ denotes the target position that the bucket's end-tip aims to reach, composed of $\hat{p}_x$, $\hat{p}_y$, and $\hat{p}_z$. Additionally, $U$ refers to the set of agent's commands, which includes command values $u_1$, $u_2$, $u_3$, and $u_4$ for each joint. Furthermore, $S$ represents a state set comprising a total of 10 dimensions, including $P$, $\hat{P}$, and $U$. Moreover, $A$ is an action set encompassing velocity values $a$ for each of the four joints. Lastly, $R$ corresponds to rewards and consists of $r$, which represents the sum of partial rewards
\subsection{Reward}
To efficiently train the 4 DOFs excavator, rewards are computed separately for each joint, and then these individual rewards are aggregated to calculate the total reward. For this purpose, distinct target values, unrelated to other joints, are established for each joint, and the rewards for each are calculated based on the difference between these target values and the current values. The reason behind setting sub-goals and computing rewards for each joint individually stems from the fact that, in contrast to the 2D case, the excavator's action space expands into 3D, presenting challenges when training a model with a single overarching goal. Therefore, we set independent targets for each joint, and computed rewards accordingly. This independent reward system is depicted in Fig. \ref{fig:rewards}. The process of calculating rewards is as follows:
\begin{gather}
    \hat{l}_{xy} = \sqrt{\left(\hat{p}_x\right)^2+\left(\hat{p}_y\right)^2},\\
    \hat{p}^{bm\cdot arm} = (\hat{l}_{xy}\cdot\sin{\theta^{swg}},\hat{l}_{xy}\cdot\cos{\theta^{swg}},\hat{p}_z+l^{bkt}),\\
    \hat{p}^{bkt} = (p^{bm\cdot arm}_x , p^{bm\cdot arm}_y , p^{bm\cdot arm}_z-l^{bkt}),\\
    d^{swg} = ||\hat{\theta}^{swg}-\theta^{swg}||_2,\label{eqn:swing_reward}\\
    d^{bm\cdot arm} = ||\hat{p}^{bm\cdot arm}-p^{bm\cdot arm}||_2,\\
    d^{bkt} = ||\hat{p}^{bkt}-p^{bkt}||_2,\\
    r^{k} = \begin{cases}
    -(d^k)^2 & \text{if }d^k > 1 \\ 4\cdot(d^k-1)^2-1 & \text otherwise \end{cases},\\
    r = c_1 \cdot r^{swg} + c_2 \cdot r^{bm\cdot arm} + r^{bkt}, 
\end{gather}
where $d^{swg}$ represents the reward for the swing joint, which corresponds to the difference between $\hat{\theta}^{swg}$ and $\theta^{swg}$. Additionally, in the left part of Fig. \ref{fig:rewards}, you can see $\hat{\theta}_{swg}$ and $\theta^{swg}$, which correspond to the target and actual angles of the swing joint along the $x$ direction. Similarly, $d^{bm\cdot arm}$ represents the difference between $p^{bm\cdot arm}$, which represents the actual position of the arm, and the target position of the arm, $\hat{p}^{bm\cdot arm}$, shown in the right part of Fig. \ref{fig:rewards}. At this point, $\hat{p}^{bm\cdot arm}$ is calculated using $\hat{p}^{bkt}$, which is the target position of the bucket's end-tip. The value $\hat{l}_{xy}$ corresponds to the distance on the $xy$ plane from the origin to $\hat{p}$. By projecting these distances into the $x$ and $y$ axes using $\theta^{swg}$, $x$ and $y$ values of $\hat{p}^{bm\cdot arm}$ are calculated. For the $z$ value of $\hat{p}^{bm\cdot arm}$, we define it as the sum of the existing $\hat{p}_z$ and the size of the bucket, $l^{bkt}$. This choice is made to maintain the bucket as perpendicular to the floor as possible during the target movement. Additionally, $d^{bkt}$ is defined as the difference between $p^{bkt}$, representing the end-tip position of the actual bucket, and $\hat{p}^{bkt}$, corresponding to the target position. Furthermore, $k$ corresponds to one of the joints ($swg, bm\cdot arm, bkt$). Moreover, $r^{swg}$, $r^{bm\cdot arm}$, and $r^{bkt}$ represent rewards applied to the swing, boom and arm, and bucket joints, respectively. These rewards are calculated differently depending on the value of $d^k$. Finally, the sum of these values is defined as $r$, representing the final reward. In this case, $c_1$ and $c_2$ are constant values used to regularize each reward.
\subsection{Learning}

We utilize the Twin Delayed Deep Deterministic policy gradient algorithm (TD3) \cite{fujimoto2018addressing}, which is an enhancement of the Deep Deterministic Policy Gradient algorithm (DDPG) \cite{lillicrap2015continuous}. As in \cite{johannink2019residual}, TD3 is chosen due to its good performance in models involving continuous action spaces, such as robot arms. The TD3 architecture is detailed in Table \ref{table:network_architecture}, comprising one action layer and two critical layers. Each layer is fully connected, and ReLU serves as the activation function.
\begin{table}[t]

\caption{Accuracy Of Point Chasing (Simulation).\\
(Euc: Euclidean distance-based reward,\\Ind: Independent reward).
}
\label{table:sim}
\begin{center}
\renewcommand{\arraystretch}{2}

    \begin{tabular}{P{0.06\textwidth} P{0.12\textwidth} P{0.06\textwidth} P{0.06\textwidth} P{0.06\textwidth} P{0.06\textwidth}}\hline

    && DDPG & \multicolumn{2}{Sc}{TD3}\\\cline{4-5}
    & &  & Euc. & Ind.\\\hline
    \multirow{2}{*}{Linear}&$\mu_{err}$ (cm)& 9.12&3.89&\textbf{2.27}\\

    &$max_{err}$ (cm)&16.22&17.18&\textbf{6.77}\\\hline
    \multirow{2}{*}{Slope}&$\mu_{err}$ (cm)& 10.76&6.21&\textbf{1.79}\\

    &$max_{err}$ (cm)&22.55&17.32&\textbf{9.67}\\\hline
    \end{tabular}

\end{center}
\end{table}
\begin{table*}[t]
\vspace{0.2cm}
\caption{Accuracy Of Point Chasing (Real Excavator).\\
(Lv: The average height level of the trajectory).}
\label{table:real}
\begin{center}
\renewcommand{\arraystretch}{2}
    \begin{tabular}{P{0.09\textwidth} P{0.14\textwidth} P{0.07\textwidth} P{0.07\textwidth} P{0.07\textwidth} P{0.07\textwidth} P{0.04\textwidth} P{0.07\textwidth} P{0.07\textwidth} P{0.07\textwidth}}\hline
    &&\multicolumn{4}{Sc}{Linear}&&\multicolumn{3}{Sc}{Slope} \\\cline{3-6}\cline{8-10}
    &&Lv0&Lv1&Lv2&Lv3&&Lv0&Lv1&Lv2\\\hline
    \multirow{2}{*}{Human}&$\mu_{err}$ (cm)&16.131&\textbf{10.584}&9.265&12.991&&14.29&13.771&14.187\\
    &$max_{err}$ (cm)&19.521&\textbf{17.253}&11.254&14.243&&24.25&\textbf{18.284}&19.472\\\hline
    
    \multirow{2}{*}{Our method}&$\mu_{err}$ (cm)&\textbf{13.913}&11.193&\textbf{11.641}&\textbf{9.808}&&\textbf{8.112}&\textbf{8.792}&\textbf{7.343}\\
                              &$max_{err}$ (cm)&\textbf{16.429}&17.657&\textbf{15.349}&\textbf{13.141}&&\textbf{18.243}&19.657&\textbf{17.481}\\\hline
    \end{tabular}
    \vspace{-0.4cm}
\end{center}
\end{table*}
\begin{figure}[t]
    \centering
    \framebox{\parbox{0.49\textwidth}{\includegraphics[width=0.49\textwidth]{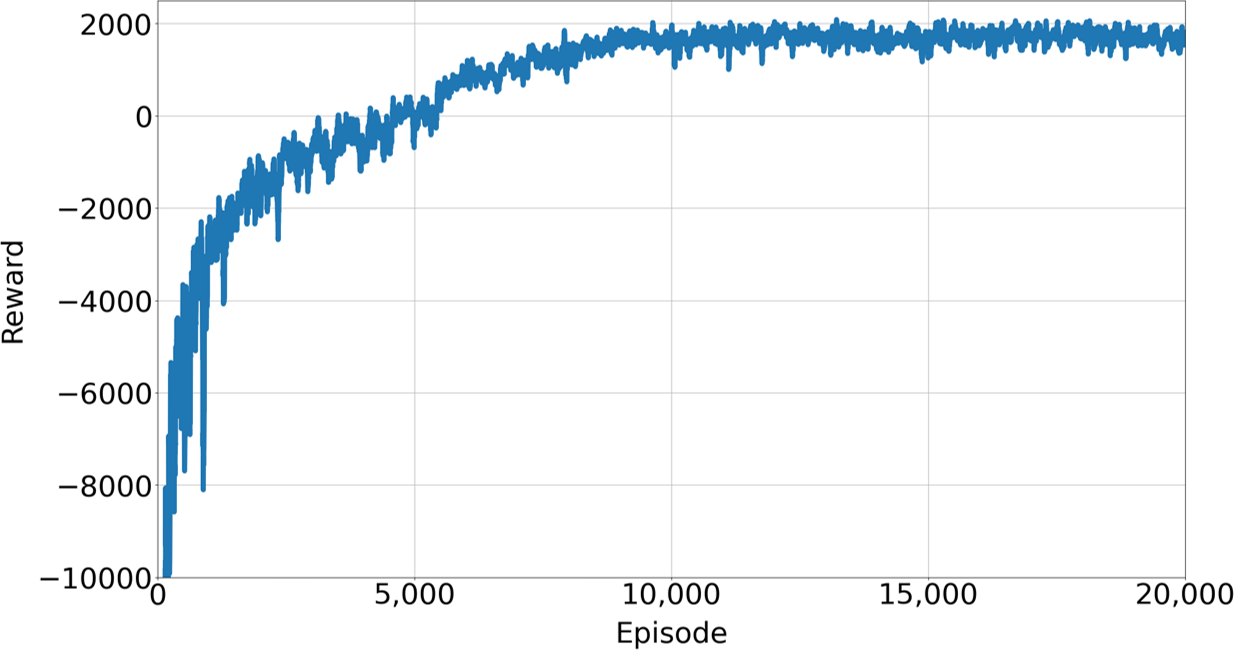}}}
    \caption{Changes in reward according to model learning.}
    \label{fig:training_graph}
        \vspace{-0.2cm}
\end{figure}
\begin{figure}[t]
    \centering
    \framebox{\parbox{0.45\textwidth}{\includegraphics[width=0.45\textwidth]{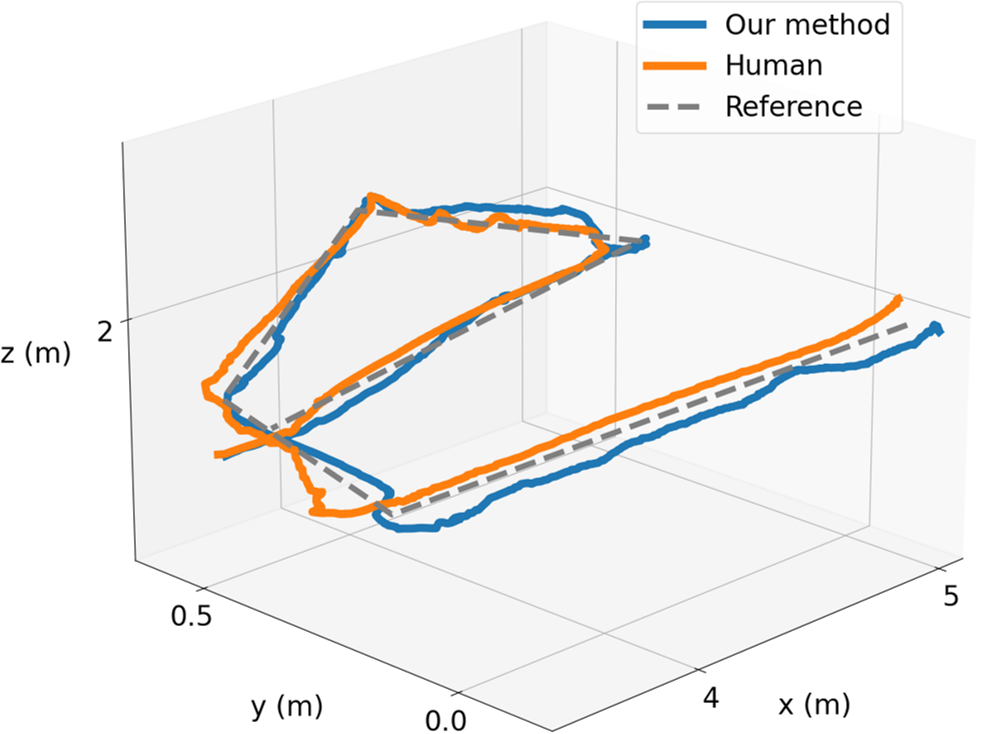}}}
    \caption{Results of point chasing in a linear trajectory.}
    \label{fig:linear_traj}
        \vspace{-0.2cm}
\end{figure}
\begin{figure}[t]
    \centering
    \framebox{\parbox{0.45\textwidth}{\includegraphics[width=0.45\textwidth]{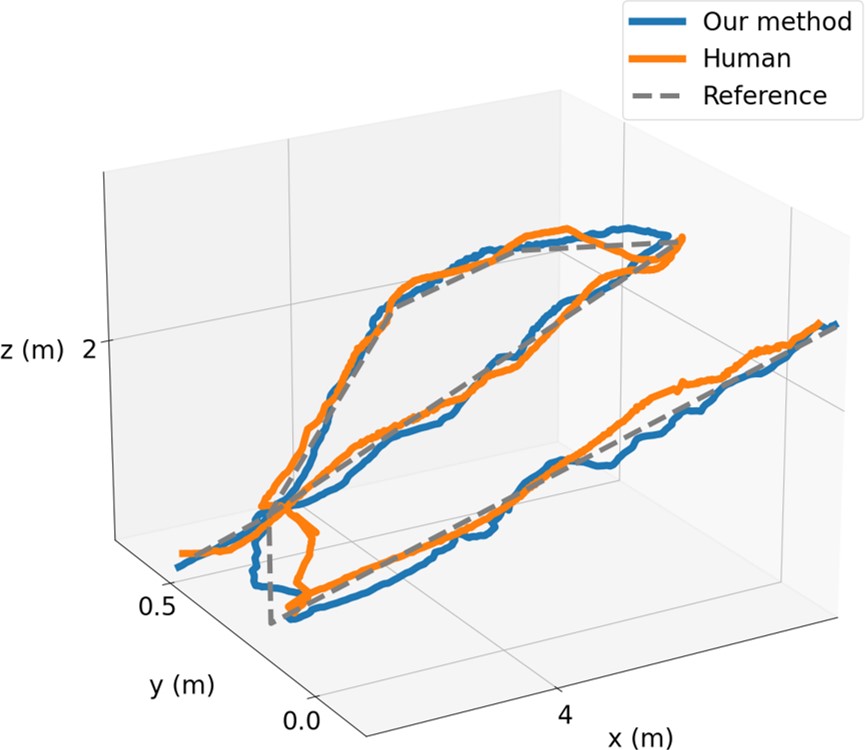}}}
    \caption{Results of point chasing in a slope trajectory.}
    \label{fig:slope_traj}
        \vspace{-0.2cm}
\end{figure}
\begin{figure}[t]
    \centering
    \vspace{0.2cm}
    \framebox{\parbox{0.45\textwidth}{\includegraphics[width=0.45\textwidth]{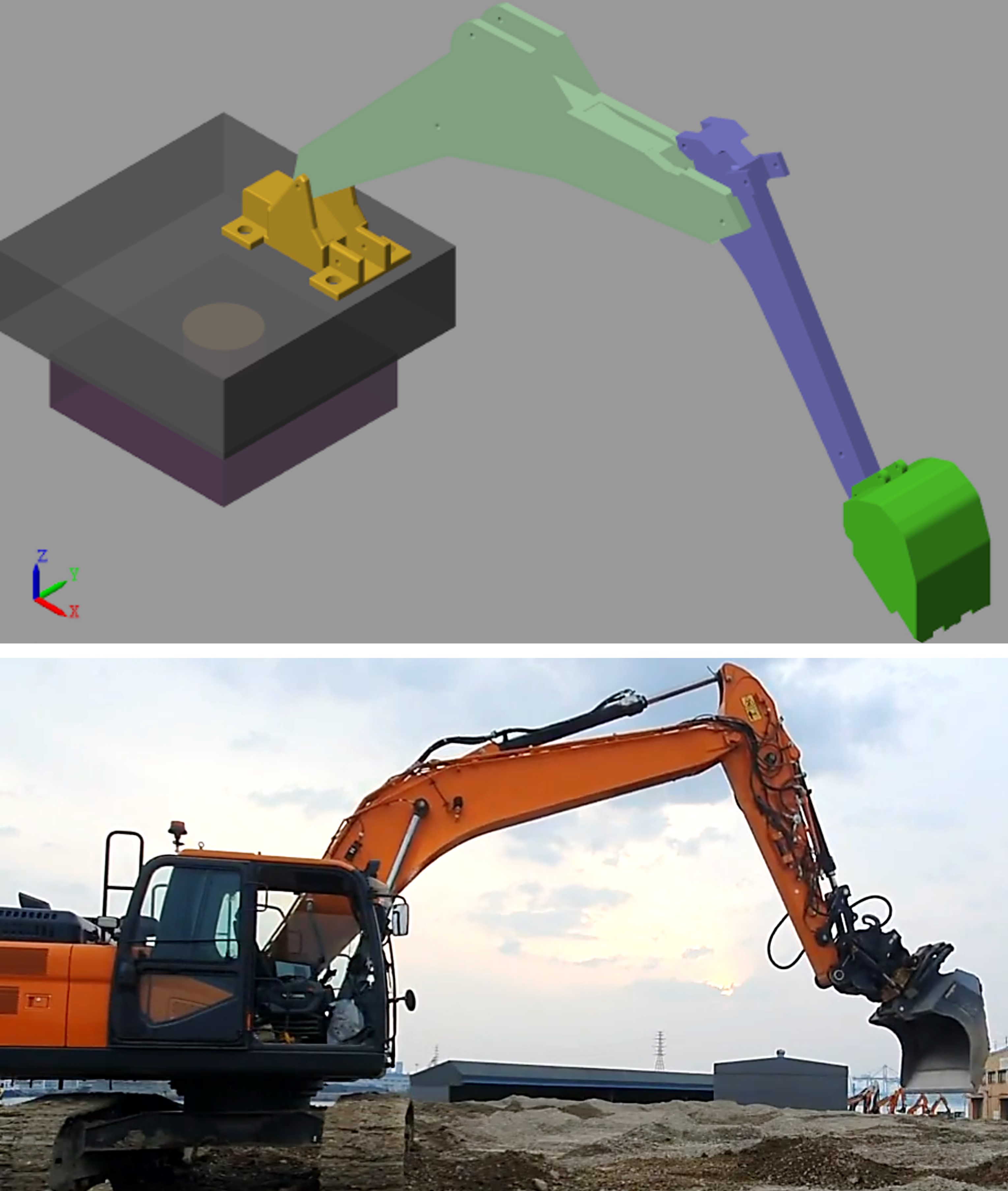}}}
    \caption{Two test environments. Simulation (Top), Real excavator (Bottom).}
    \label{fig:environments}
    \vspace{-0.2cm}
\end{figure}
\section{Experiments}
\subsection{Settings}
In this paper, we conducted a performance comparison among different reinforcement learning methods, selecting TD3 and DDPG as the methods of comparison. As part of an ablation study, we also evaluated the performance improvement of the TD3 algorithm in a simulation environment using an alternative Euclidean distance-based reward, which measures the distance between the bucket's end-tip position and the target position, instead of the independent reward proposed in this paper. The evaluation of the algorithms involved calculating the Root Mean Square Error (RMSE) for the Euclidean distance between the target point and the bucket end, measured in meters. The values $\mu_{err}$ and $max_{err}$ in Table \ref{table:sim} and \ref{table:real} represent the mean and maximum errors, respectively, in each trajectory. Moreover, we verified the performance through two tasks using reinforcement learning: Linear and Slope. First, in the case of linear flat work, it is the most basic and frequent task for excavator workers, involving the spreading and flattening of uneven or large piles of soil. We pre-defined the trajectory for this flattening operation and made the bucket follow the trajectory. Additionally, in the case of the second slope flattening task, it is almost similar to the previous linear flattening work, but it is not parallel to the floor; instead, it is inclined.
\subsection{Train}
$20,000$ episodes were applied consistently across all methods. Each episode was designed to progress to the next when the number of steps exceeded the predefined length or when they were terminated prematurely. At this point, the previously defined length was set at $2,048$. The corresponding results are shown in Fig. \ref{fig:training_graph}. It can be observed that, up to $10,000$ episodes, the reward gradually increases with fluctuations and converges to $1,500$. Additionally, during model learning, a learning rate of $10^{-5}$ was utilized. Furthermore, a noise variation of 0.5 was employed. In the case of the target position, $P^{tgt}$ used in point chasing during learning, random sampling was conducted within a certain range to reflect the actual working environment of the fork lane. For instance, $p^{tgt}_x$ was chosen to be between 2 and 5, while $p^{tgt}_z$ was selected as a value greater than -1. All units of the target position were set in meters.
\subsection{Simulation}
As in Fig. \ref{fig:environments}, simulations were conducted using \emph{Simulink\footnote{https://www.mathworks.com/products/simulink.html}}. A simulation environment closely resembling a real excavator setup was established by importing modeling files into \emph{Simulink} and defining articulation joints for each link. Furthermore, parts of the pre-defined trajectory for various tasks (linear and slope) were sequentially invoked over time to set target locations, thereby configuring an environment in which the excavator model could perform the point-chasing method. Experimental results in this configured simulation environment, as shown in Table \ref{table:sim}, indicated that TD3 outperformed other algorithms in all tasks. Additionally, among the rewards tested with the TD3, independent rewards exhibited better performance than Euclidean distance-based rewards. Consequently, we can conclude that independent rewards are more suitable for model training than single Euclidean distance-based rewards, as they allocate better sub-targets for each joint, as evidenced by their superior performance.
\subsection{Real Environment}
For the real vehicle experiments, a Hyundai-InfraCore DX225LC-7 model excavator was employed, as shown in Fig. \ref{fig:environments}. The excavator was secured to the ground, and only the articulated parts of the vehicle were allowed to move for conducting the experiments. To calculate the bucket's endpoint position during the actual vehicle experiments, IMU sensors and a swing angle sensor were installed on each joint, as shown in Fig. \ref{fig:process}. The real-time position of the bucket was estimated by integrating the values of these sensors for each joint based on kinematics. The detailed position estimation method is described in the paper \cite{sun2020sensor}. The model utilized for the actual excavator experiments was trained using the TD3 with the independent reward proposed in our paper, which demonstrated the highest performance in the previous simulation experiment. The model's performance was verified for a single case using the proposed method and compared against the performance of a skilled operator conducting similar tasks. The results, as presented in Table \ref{table:real}, indicated that our proposed approach outperformed the skilled operator in most tasks. These results were consistent across both linear and slope trajectory outcomes, as illustrated in Fig. \ref{fig:linear_traj} and \ref{fig:slope_traj}. For reference, we tested 3 to 4 trajectories with varying average heights for each task, as indicated by Lv in Table \ref{table:real}. Fig. \ref{fig:linear_traj} and \ref{fig:slope_traj} depict the case of Lv0, indicating the lowest average height among these trajectories. Therefore, we conclude that when training using the reinforcement learning method based on independent rewards, our approach achieves high accuracy in performing tasks in both simulation and real excavator environments.
\section{Conclusion}
In this paper, we propose a 3D control method for autonomous excavators using independent reward-based reinforcement learning. A trained agent commands the operation of each joint of the excavator to move the bucket's endpoint to the target position, enabling the excavator to perform the desired tasks continuously. We construct a 4 DOFs model, including swing, boom, arm, and bucket joints. Among these, the cabin swing joint, which has received less attention in previous works, is a key target for control. Cabin swing rotation allows lateral movement of the bucket, expanding the bucket's operational radius from the conventional 2D plane to a 3D space. As a result, this expanded action space significantly increases the practical applicability in construction sites. However, the inclusion of the swing joint as a control target results in an expanded workspace of the excavator. Consequently, due to the significant changes in the bucket's endpoint caused by swing joint values, in a single-target learning, the learning of swing joint control may lag behind the rest of the joint learning, hampering overall progress. To solve this, we enhance the model's learning by setting separate targets for each joint during the training. This allows each joint to learn independently regardless of the progress of other joints, resulting in faster and more accurate model training. Additionally, to verify the model's performance, tests were conducted in both virtual and real excavator environments. The validation confirms that our model operates with high accuracy in both environments. In conclusion, the proposed model performs well in 4 DOFs settings and excels in performing two different tasks (linear and slope), as demonstrated through extensive testing.
\section*{Acknowledgement}
This work is supported by the National Research Foundation of Korea (NRF) through the Ministry of Science and ICT under Grant  2021R1A2C1093957, by Korean Ministry of Land, Infrastructure and Transport(MOLIT) as Innovative Talent Education Program for Smart City, and by Korea Institute for Advancement of Technology(KIAT) grant funded by the Korea Government(MOTIE) (P0020536, HRD Program for Industrial Innovation). The Institute of Engineering Research at Seoul National University provided research facilities for this work.
\bibliographystyle{IEEEtran}
\bibliography{root}
\end{document}